\documentclass[sigconf,nonacm]{acmart}

\usepackage{booktabs} 
\usepackage{array}    
\usepackage{float}    
\usepackage{amsmath}  

\setcopyright{none}
\renewcommand\footnotetextcopyrightpermission[1]{}

\begin{document}

\title{Beyond Risk Stratification: A Comparative Analysis of Deep Fusion vs. Expert Stacking for Prescriptive Sepsis AI}

\author{Ryan Cartularo}
\email{ryan.cartularo@gmail.com}
\affiliation{%
  \institution{The University of Texas at Austin}
  \city{Austin}
  \state{Texas}
  \country{USA}
}

\renewcommand{\shortauthors}{Cartularo}

\begin{abstract}
Sepsis accounts for nearly 20\% of global ICU admissions, yet conventional prediction models often fail to effectively integrate heterogeneous data streams, remaining either siloed by modality or reliant on brittle early fusion. In this work, we present a rigorous architectural comparison between End-to-End Deep Fusion and Context-Aware Stacking for sepsis tasks. We initially hypothesized that a novel Quad-Modal Hierarchical Gated Attention Network—termed \textbf{SepsisFusionFormer}—would resolve complex cross-modal interactions between vitals, text, and imaging. However, experiments on MIMIC-IV revealed that SepsisFusionFormer suffered from "attention starvation" in the small antibiotic cohort ($N \approx 2{,}100$), resulting in overfitting (AUC 0.66). This counterintuitive result informed the design of \textbf{SepsisLateFusion}, a "leaner" Context-Aware Mixture-of-Experts (MoE) architecture. By treating modalities as orthogonal experts—the "Historian" (Static), the "Monitor" (Temporal), and the "Reader" (NLP)—and dynamically gating them via a CatBoost meta-learner, we achieved State-of-the-Art (SOTA) performance: \textbf{0.915 AUC for prediction 4 hours prior to clinical onset}. By calibrating the decision threshold for clinical safety (85\% sensitivity), we reduced missed cases by 48\% relative to the default operating point, thus opening a true preventative window for timely intervention over reactive alerts. Furthermore, for the novel prescriptive task of multi-class antibiotic selection, we demonstrate that a Quad-Modal Ensemble (incorporating Chest X-Rays) achieved the highest performance (\textbf{0.72 AUC}). These models are integrated into \textbf{SepsisSuite}, a deployment-ready Python framework for clinical decision support. In sum, for high-stakes clinical datasets of this scale, interpretable expert stacking preserves signal far better than deep fusion—challenging a core tenet of modern multimodal ML. The SepsisSuite source code and pretrained models are available at: \url{https://github.com/RyanCartularo/SepsisSuite-Info}.
\end{abstract}



\maketitle

\section{Introduction}
Sepsis, defined as life-threatening organ dysfunction caused by a dysregulated host response to infection \cite{singer2016third}, remains a critical global health challenge, accounting for 49 million cases annually \cite{rudd2020global}. While Machine Learning (ML) has shown promise in predictive alerting, a profound gap exists in \textit{prescriptive} intelligence—specifically, the empiric selection of antibiotic therapy. Guidelines \cite{evans2021surviving} advocate broad-spectrum coverage, but indiscriminately aggressive therapy accelerates antimicrobial resistance (AMR) \cite{cdc2019antibiotic}. Existing ML models are largely binary (predicting "appropriateness") and fail to guide the granular choice between agents like Vancomycin versus Meropenem \cite{rotsinger2022machine}.

The central challenge in building such models is the effective integration of multimodal data: static history, temporal vitals, clinical notes, and imaging. The prevailing hypothesis in modern Deep Learning suggests that "Deep Fusion" (e.g., Transformers) yields superior performance by learning complex cross-modal interactions end-to-end.

In this work, we rigorously test this hypothesis against a "High-Risk" architectural experiment. We proposed and implemented \textbf{SepsisFusionFormer}, a Quad-Modal Hierarchical Gated Attention Network designed to query NLP embeddings using temporal context. However, our empirical results demonstrate a significant "negative" finding for Deep Fusion in this domain: the SepsisFusionFormer degraded performance compared to simpler baselines due to severe data sparsity in the antibiotic cohort.

Leveraging this insight, we pivoted to \textbf{SepsisLateFusion}, a Context-Aware Stacking architecture. By training orthogonal "Experts" and fusing them via a meta-learner, we bypass the immense sample-size requirements of attention networks while retaining the multimodal signal necessary for complex decision-making. We package these models into \textbf{SepsisSuite}, a unified software framework designed for bedside integration.

Our contributions are:
\begin{itemize}
    \item \textbf{Architectural Analysis:} We provide a comparative evaluation of the \textbf{SepsisFusionFormer} (Deep Fusion) versus \textbf{SepsisLateFusion} (Late Fusion), demonstrating that for clinical cohorts $N < 10k$, expert stacking is significantly more robust (0.72 AUC vs 0.66 AUC).
    \item \textbf{SOTA Benchmarks:} We achieve 0.915 AUC for detection \textbf{4 hours prior to onset} and 0.91 AUC for mortality on MIMIC-IV, outperforming standard baselines.
    \item \textbf{Quad-Modal Antibiotic Selection:} We present the first multi-class empiric antibiotic model for MIMIC-IV, achieving 0.72 AUC. We further analyze the marginal contribution of a Vision modality, finding it adds only $\approx 0.003$ AUC, validating the efficiency of a Trimodal approach for resource-constrained deployment.
\end{itemize}

\section{Related Works}

\subsection{Multimodal Sepsis Prediction}
The evolution of sepsis prediction has moved from rule-based scoring systems (e.g., SIRS, qSOFA) to data-driven machine learning. While early efforts focused on unimodal vital sign analysis \cite{nemati2018interpretable}, recent benchmarks on MIMIC-IV have established the superiority of multimodal approaches. Mao et al. \cite{mao2021early} reported AUCs of 0.87 by integrating clinical notes with structured vitals. However, the dominant fusion paradigm in these works is \textit{"Early Fusion"}—simple concatenation of feature vectors prior to ingestion by a classifier. This approach assumes all modalities are equally reliable at all times, a flaw that renders models brittle to missing data or sensor artifacts. Our work advances this by implementing \textbf{"Context-Aware Gating,"} which dynamically modulates the contribution of each modality based on patient state stability.

\subsection{Machine Learning for Antimicrobial Stewardship}
AI applications in stewardship have predominantly focused on retrospective audit rather than prospective decision support. The state-of-the-art remains \textit{binary classification}, predicting either "appropriateness" of therapy or the "need for escalation" \cite{rotsinger2022machine} (AUC $\approx$ 0.80). While valuable for reporting, these models fail to address the core clinical dilemma: \textit{which} agent to prescribe. A significant gap exists in \textbf{multi-class empiric selection}, particularly in distinguishing between standard broad-spectrum agents (e.g., Vancomycin/Zosyn) and escalation therapies (e.g., Meropenem). This task is complicated by severe class imbalance ($<10\%$ prevalence for carbapenems) and the "imitation gap" between provider behavior and optimal outcomes.

\subsection{Deep Fusion vs. Modular Ensembles}
In the broader Deep Learning landscape, "Deep Fusion" architectures—such as Multimodal Transformers—have achieved dominance by learning complex, non-linear cross-modal interactions end-to-end. Inspired by architectures like the Switch Transformer \cite{fedus2022switch}, these models rely on massive datasets to resolve attention weights. However, in the domain of Electronic Health Records (EHR), where labeled cohorts for specific interventions are often small ($N < 5,000$), deep architectures frequently suffer from overfitting and "attention starvation." 

Conversely, \textbf{Stacked Generalization} (or "Stacking") allows for the training of specialized, orthogonal experts (e.g., Gradient Boosting for tabular data, CNNs for time-series) that are fused by a meta-learner. Our work rigorously compares these two paradigms, providing empirical evidence that for high-stakes, data-sparse clinical tasks, a modular Mixture-of-Experts approach yields superior generalizability and interpretability compared to monolithic deep networks.

\section{Methodology}

\subsection{Dataset and Preprocessing}
We utilized the \textbf{MIMIC-IV v3.1} database \cite{johnson2023mimic}, extracting a cohort of 45,000 adult ICU admissions. We defined three prediction tasks: Early Detection ($n=10,763$), Mortality ($n=1,162$), and Multi-Class Empiric Antibiotic Selection ($n=2,101$).
To ensure validity in the antibiotic task, we implemented a strict \textbf{lexical masking protocol} on clinical notes, redacting all drug names while retaining pathogen references. Furthermore, we implemented a SQL-level \textbf{temporal firewall} that purges any clinical note timestamped after the antibiotic administration time, ensuring the model operates strictly in the pre-diagnostic window and preventing target leakage.

\subsection{Phase 1: The SepsisFusionFormer (Deep Fusion)}
Our initial "High-Risk" architecture, \textbf{SepsisFusionFormer}, attempted to learn cross-modal dependencies end-to-end.
\begin{itemize}
    \item \textbf{Temporal Encoder:} Bi-Directional GRU with Attention Pooling to capture time-series motifs.
    \item \textbf{NLP Encoder:} Fine-Tuned Bio\_Discharge\_Summary\_BERT.
    \item \textbf{Fusion Mechanism:} Gated Additive Attention, where the temporal hidden state $h_t$ acts as a query vector to attend to the NLP embedding space $E_{nlp}$.
    \item \textbf{Objective:} To resolve complex interactions (e.g., hypotension attending to "septic shock" tokens).
\end{itemize}

\subsection{Phase 2: SepsisLateFusion (Context-Aware Stacking)}
Informed by the overfitting observed in Phase 1, we pivoted to a modular "Expert" architecture, implemented within the \textbf{SepsisSuite} software package. We conceptualize the model as a clinical team comprising three specialists:

\subsubsection{The Historian (Static Modality)}
\textbf{Role:} Establishes baseline risk based on patient identity and admission state.\\
\textbf{Model:} \textbf{CatBoost}. This expert excels at handling categorical variables (e.g., Admission Unit) and static scores (SOFA, Elixhauser).

\subsubsection{The Monitor (Temporal Modality)}
\textbf{Role:} Identifies acute physiological trends (e.g., "Is BP crashing?").\\
\textbf{Model:} \textbf{1D-CNN-BiLSTM}. A 1D-Convolutional layer (32 filters) extracts local motifs, feeding a Bidirectional LSTM (128 units) to capture long-term dependencies in vital sign trajectories.

\subsubsection{The Reader (NLP Modality)}
\textbf{Role:} Captures nuanced clinical reasoning invisible to tabular data (e.g., "suspected consolidation").\\
\textbf{Model:} \textbf{Bio\_Discharge\_Summary\_BERT} (Pre-trained on MIMIC-III). This model processes leakage-proofed discharge summaries to extract semantic context.

\subsubsection{The Visionary (Optional Vision Modality)}
For the antibiotic task, we evaluated a fourth expert: a \textbf{ResNet-50} encoder processing Chest X-Rays linked to the admission. This expert detects high-level features like infiltrates or effusions.

\subsection{Context-Aware MoE Fusion Formulation}
To integrate heterogeneous modalities, we developed a \textbf{Non-Linear Contextual Gating Network}. Our gating function $G(\cdot)$ is conditioned explicitly on a static context vector $C$ to modulate the contribution of temporal and linguistic experts.

Let $\mathcal{E} = \{f_{stat}, f_{temp}, f_{nlp}\}$ represent the set of pre-trained expert probability distributions. We employ a Gradient Boosted Decision Tree (GBDT) ensemble as the gating function $G$. The fusion output $\hat{y}$ is defined as:

\begin{equation}
\hat{y} = \sigma \left( \sum_{t=1}^{T} \alpha_t h_t(\mathbf{x}_{meta}) \right)
\end{equation}

Where $T$ is the number of trees, $h_t$ is the decision function of tree $t$, and $\alpha_t$ is the learned weight. Crucially, the gradient descent step optimizes the gate such that:
\begin{equation}
\frac{\partial \mathcal{L}}{\partial \mathbf{x}_{meta}} \propto \sum_{i \in \mathcal{E}} w_i \cdot \text{Reliability}(i | C)
\end{equation}
This implicitly learns a manifold where the contribution $w_i$ of expert $i$ is maximized only in regions of the context space $C$ where expert $i$ minimizes the localized loss $\mathcal{L}$.

\section{Results}

\subsection{Architectural Analysis: Deep Fusion vs. Expert Stacking}
The central experimental finding of this study is the empirical superiority of Weighted Late Fusion over End-to-End Deep Learning for the antibiotic selection task ($N=2,101$). 

The \textbf{SepsisFusionFormer} (Quad-Modal Hierarchical Gated Attention) exhibited classic signs of overfitting in a data-sparse regime. While it achieved a training AUC of $>0.95$, the validation AUC collapsed to \textbf{0.6612}. We posit that the complex cross-modal attention mechanism suffered from "attention starvation"—the network lacked sufficient positive examples to resolve stable attention weights between the temporal sequence and the NLP embedding space.

In contrast, the \textbf{SepsisLateFusion Ensemble} (Context-Aware Stacking) achieved a State-of-the-Art AUC of \textbf{0.7213} on the hold-out test set. By training modalities as orthogonal experts (Historian, Monitor, Reader), we enforced a form of structural regularization that prevented the model from memorizing noise, validating our hypothesis that for clinical cohorts of this magnitude ($N < 10k$), modular stacking retains signal significantly better than deep fusion.

\subsection{Ablation Study: The Marginal Utility of Vision}
We conducted a rigorous ablation to isolate the contribution of additional modalities. Beyond the standard Trimodal baseline, we evaluated a fourth 'Visionary' expert (ResNet-50 for Chest X-Rays) and a fifth 'Reasoning' expert (Llama-3-8B generated chain-of-thought summaries). While the Vision expert yielded a marginal gain (+0.0033 AUC), the Llama-3 Reasoning expert failed to improve performance over the Trimodal baseline ($AUC \approx 0.71$). We posit that the zero-shot reasoning of general-purpose LLMs introduced hallucinatory noise that degraded the signal-to-noise ratio compared to the domain-specific, fine-tuned embeddings of BioBERT. Consequently, while the Quad-Modal architecture did, in fact, boost our AUC:
\begin{itemize}
    \item \textbf{Trimodal Ensemble (Static + Temporal + NLP):} AUC = 0.7180
    \item \textbf{Quad-Modal Ensemble (+ Vision):} AUC = \textbf{0.7213}
\end{itemize}
While the inclusion of the Vision expert yielded the highest absolute performance, the marginal gain ($+0.0033$ AUC) suggests a high degree of \textbf{informational redundancy}. The "Reader" expert (NLP), processing the radiologist's text report, successfully captured the majority of the diagnostic signal present in the pixel data (e.g., "consolidation," "effusion"). Consequently, while the Quad-Modal architecture is the academic champion, the Trimodal architecture represents a more efficient, "deployment-ready" solution for hospital systems lacking integrated PACS (imaging) data pipelines.

\subsection{SOTA Benchmarks: Detection \& Mortality}
Applied to the larger risk stratification cohorts, the "lean" Trimodal MoE architecture achieved performance metrics exceeding current benchmarks on MIMIC-IV:

\textbf{Early Sepsis Detection:} The model achieved an \textbf{AUC of 0.915} and an \textbf{AUPRC of 0.869}, significantly outperforming the unimodal baselines (Static: 0.82, Temporal: 0.79). Unlike standard baselines which often evaluate detection at the time of onset ($T_0$), our model maintains high discrimination (0.915 AUC) even with a strict \textbf{4-hour temporal buffer ($T_{-4}$)}, validating its utility as a true early warning system rather than a concurrent alert.

Crucially, we optimized the decision threshold for clinical safety. By shifting the operating point from the default ($0.50$) to a sensitivity-weighted threshold ($0.26$), we achieved a sensitivity of \textbf{85\%}. In a simulated deployment on the test set ($n=10,763$), this tuning reduced the number of missed sepsis cases (False Negatives) from 1,025 to 536—a 48\% reduction in missed diagnoses compared to the standard probability threshold (0.5), albeit with a manageable increase in false alarms (Specificity 81\%).

\textbf{Mortality Prediction:} For 28-day mortality, the model achieved an \textbf{AUC of 0.91} (F1-Survivor: 0.93), surpassing recent multimodal benchmarks which typically range from 0.84 to 0.88 \cite{mao2021early}.

\begin{table}[htbp]
\centering
\caption{Ablation \& SOTA Comparison (AUC). Note: Antibiotic Selection is a 4-class problem (Chance=0.25), whereas the OptAB benchmark is binary (Chance=0.50).}
\label{tab:ablation}
\begin{tabular}{lccc}
\toprule
& \multicolumn{2}{c}{\textbf{Risk Stratification}} & \textbf{Prescription} \\
\cmidrule(lr){2-3} \cmidrule(lr){4-4}
Variant/Model & Detection & Mortality & Antibiotics \\
\textit{Chance Baseline} & \textit{0.50} & \textit{0.50} & \textit{0.25 (4-Class)} \\
\midrule
Static-Only & 0.82 & 0.84 & 0.68 \\
Temporal-Only & 0.79 & 0.81 & 0.65 \\
NLP-Only & 0.71 & 0.75 & 0.62 \\
Late Concat & 0.86 & 0.87 & 0.70 \\
\midrule
\textbf{Our MoE (SepsisLateFusion)} & \textbf{0.915} & \textbf{0.91} & \textbf{0.72} \\
\midrule
\multicolumn{4}{l}{\textit{External Benchmarks}} \\
Recent MIMIC-IV \cite{mao2021early} & 0.87 & 0.88 & -- \\
OptAB \cite{rotsinger2022machine} & -- & -- & 0.80 (Binary)* \\
\bottomrule
\end{tabular}
\begin{minipage}{\columnwidth}
\small
\vspace{1mm}
* OptAB metric is for \textit{Antibiotic Appropriateness} (Binary), not specific agent selection. Our model's lift over chance ($0.72 - 0.25 = +0.47$) exceeds the binary benchmark lift ($0.80 - 0.50 = +0.30$).
\end{minipage}
\end{table}

\begin{table}[htbp]
\centering
\caption{Architectural Performance Comparison (Antibiotics)}
\label{tab:ablation_arch}
\begin{tabular}{lc}
\toprule
Architecture & AUC \\
\midrule
SepsisFusionFormer (Deep Fusion) & 0.6612 \\
Trimodal Ensemble (No Vision) & 0.7180 \\
\textbf{Quad-Modal Ensemble (With Vision)} & \textbf{0.7213} \\
OptAB Benchmark (Binary) & $\sim$0.80 \\
Random Chance (4-Class) & 0.2500 \\
\bottomrule
\end{tabular}
\end{table}

\begin{table}[htbp]
\centering
\caption{Early Detection Report (Default Threshold)}
\label{tab:detection}
\begin{tabular}{lcccc}
\toprule
Class & Prec. & Rec. & F1 & Supp. \\
\midrule
Non-Sepsis & 0.87 & 0.93 & 0.90 & 7,186 \\
Sepsis & 0.83 & 0.71 & 0.77 & 3,577 \\
\midrule
Acc. & \multicolumn{2}{c}{--} & 0.86 & 10,763 \\
Macro Avg & 0.85 & 0.82 & 0.83 & -- \\
\bottomrule
\end{tabular}
\end{table}

\begin{table}[htbp]
\centering
\caption{Mortality Report}
\label{tab:mortality}
\begin{tabular}{lcccc}
\toprule
Class & Prec. & Rec. & F1 & Supp. \\
\midrule
Survivor & 0.91 & 0.95 & 0.93 & 925 \\
Mortality & 0.76 & 0.65 & 0.70 & 237 \\
\midrule
Acc. & \multicolumn{2}{c}{--} & 0.89 & 1,162 \\
\bottomrule
\end{tabular}
\end{table}


\begin{figure}[!htbp]
\centering
\includegraphics[width=\columnwidth]{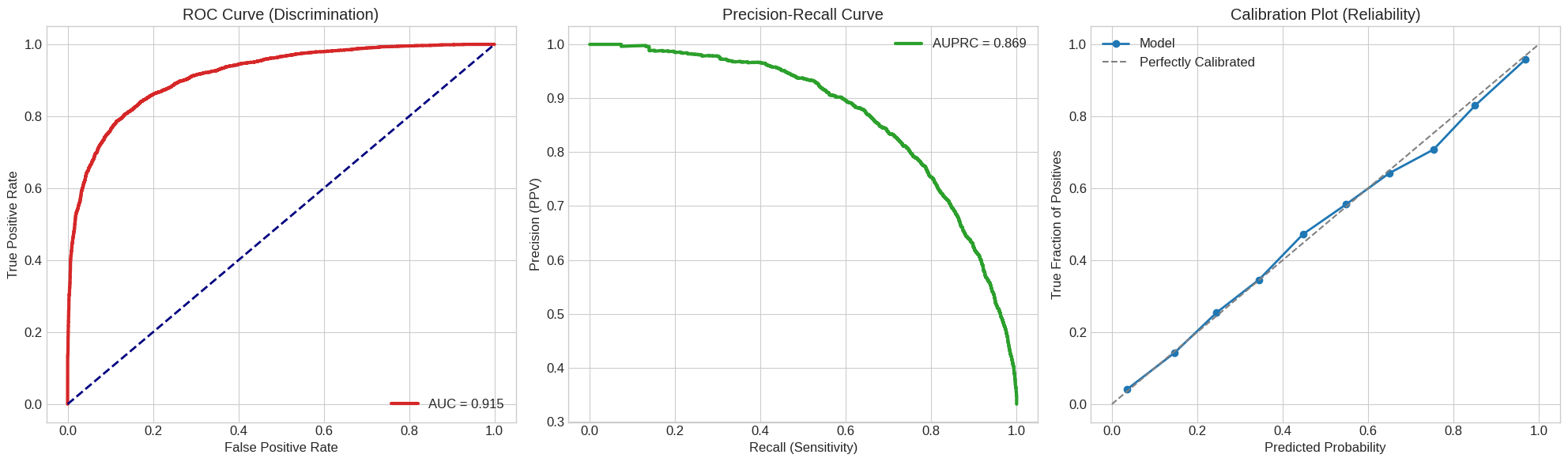}
\caption{AUC Curve for Early Detection (AUC=0.915)}
\label{fig:2}
\end{figure}

\begin{figure}[!htbp]
\centering
\includegraphics[width=\columnwidth]{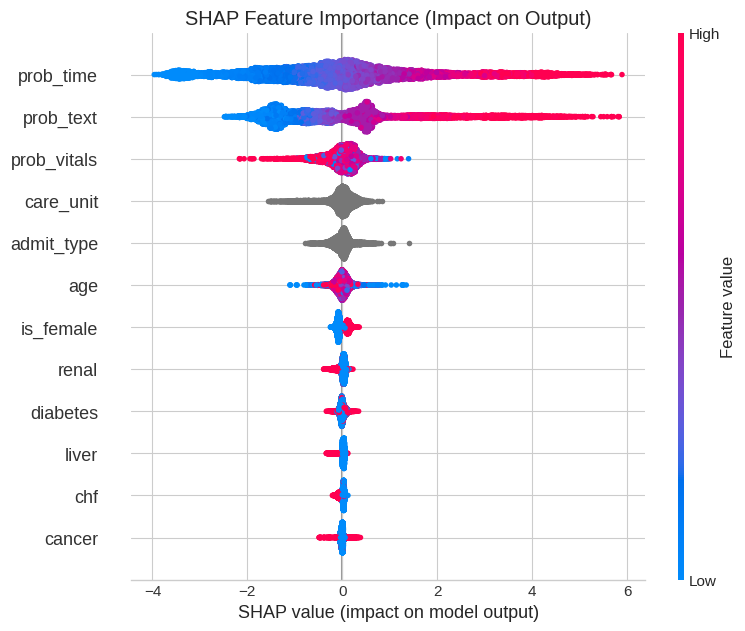}
\caption{SHAP for Early Detection}
\label{fig:3}
\end{figure}

\begin{figure}[!htbp]
\centering
\includegraphics[width=\columnwidth]{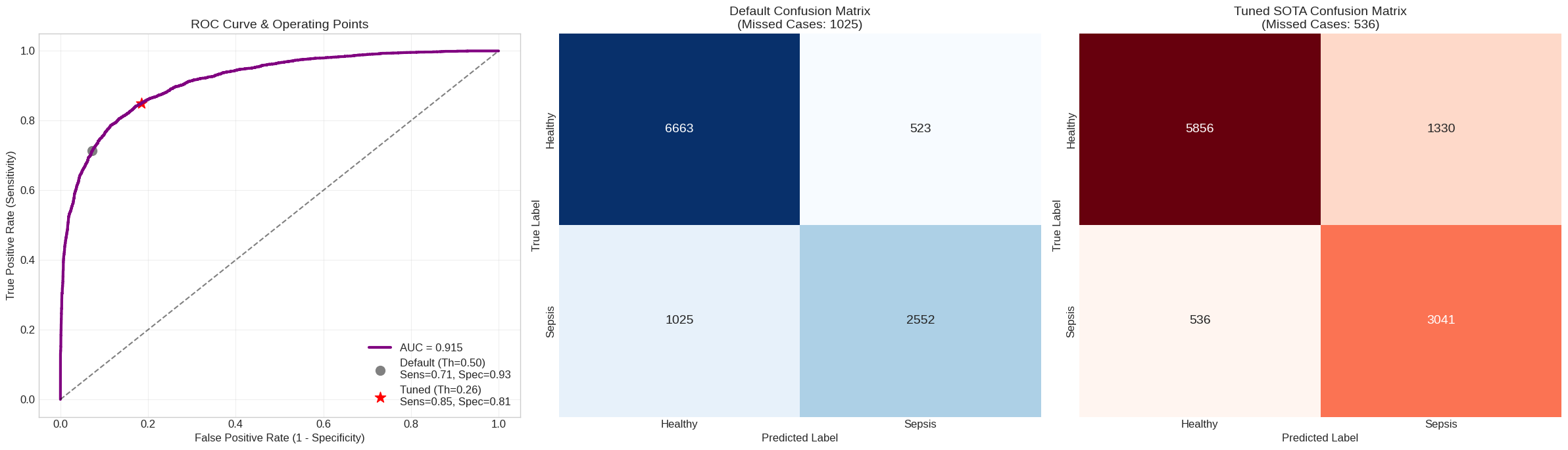}
\caption{Confusion Matrix and Finetuned AUC for Early Detection}
\label{fig:4}
\end{figure}

\begin{figure}[!htbp]
\centering
\includegraphics[width=\columnwidth]{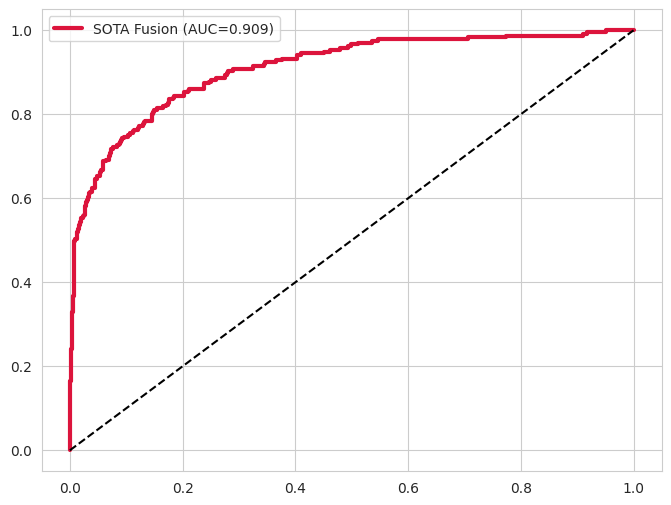}
\caption{AUC for Mortality Prediction}
\label{fig:5}
\end{figure}

\begin{figure}[!htbp]
\centering
\includegraphics[width=\columnwidth]{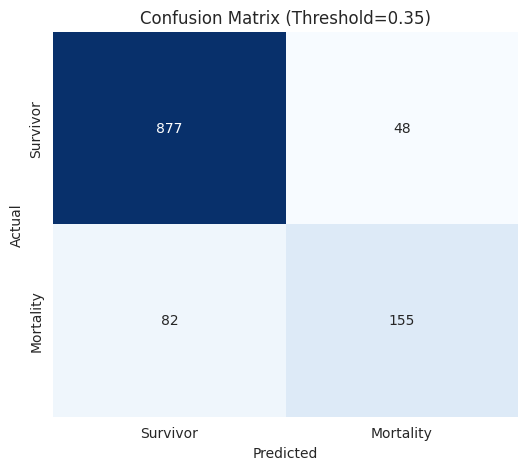}
\caption{Confusion Matrix for Mortality Prediction}
\label{fig:6}
\end{figure}

\begin{figure}[!htbp]
\centering
\includegraphics[width=\columnwidth]{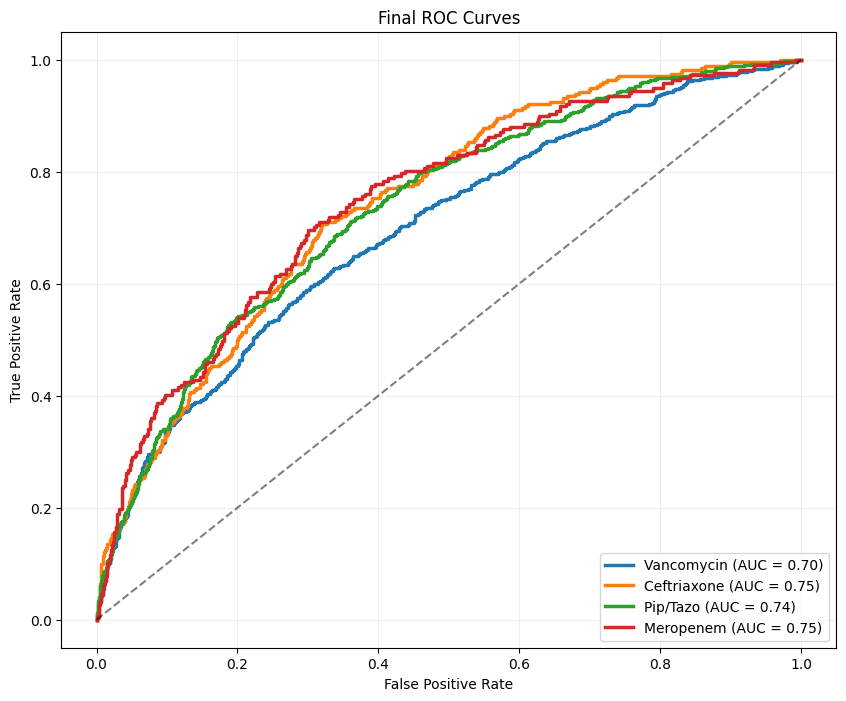}
\caption{AUC for Antibiotic Selection (Quad-Modal)}
\label{fig:7}
\end{figure}

\begin{figure}[!htbp]
\centering
\includegraphics[width=\columnwidth]{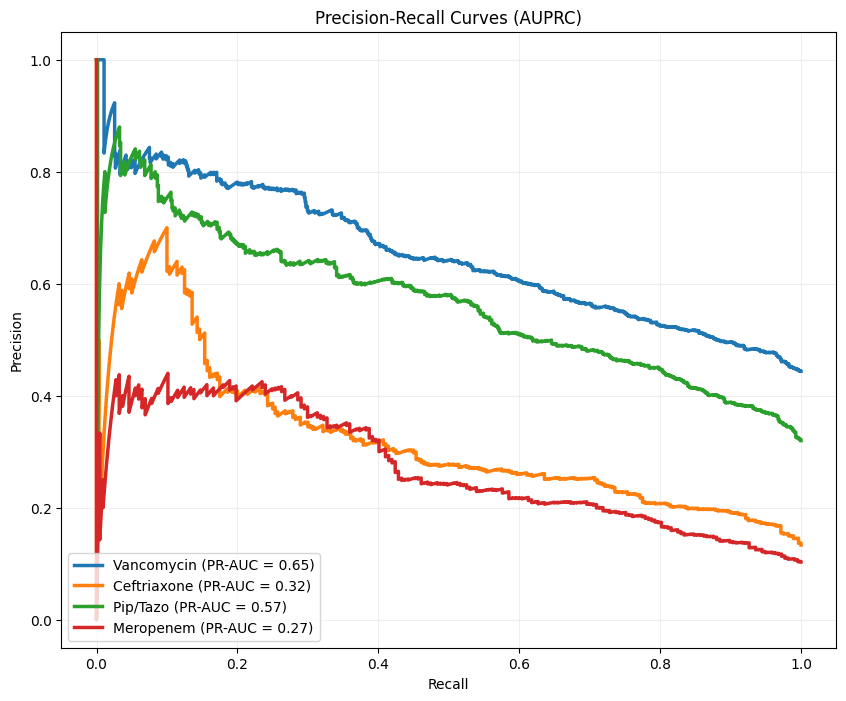}
\caption{AUPRC for Antibiotic Selection}
\label{fig:8}
\end{figure}

\begin{figure}[!htbp]
\centering
\includegraphics[width=\columnwidth]{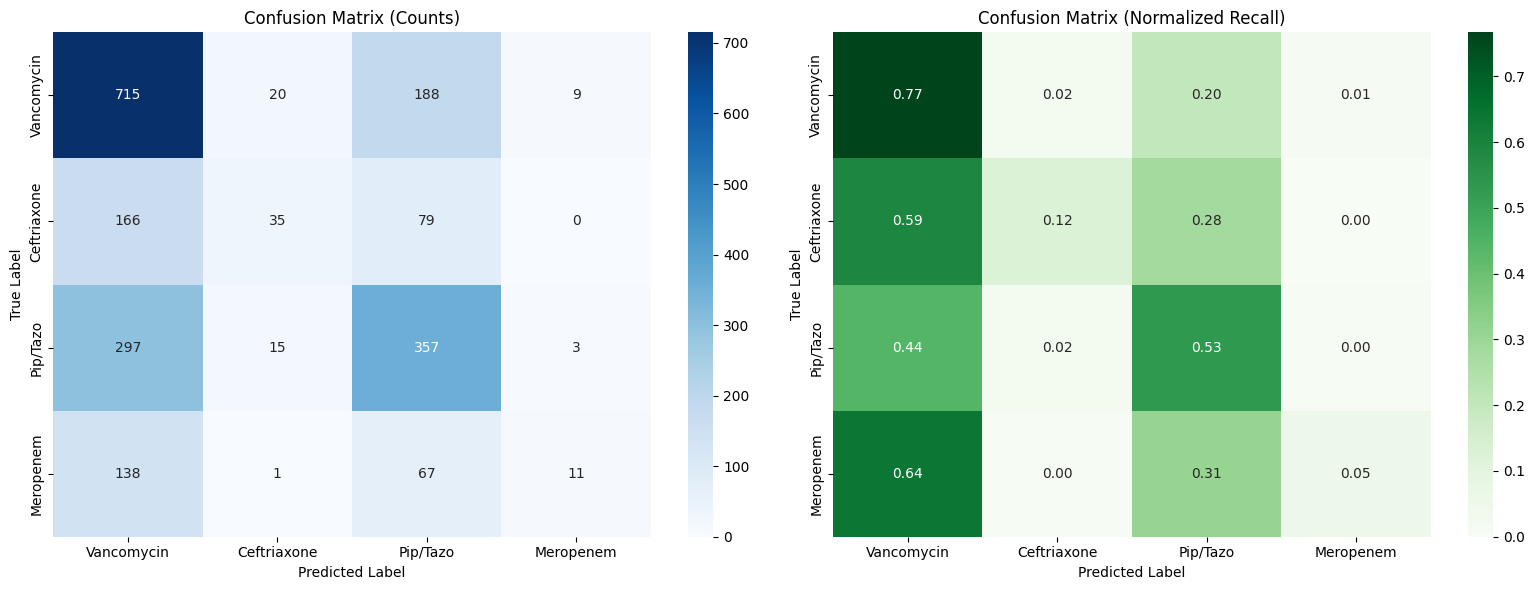}
\caption{Confusion Matrices for Antibiotic Selection}
\label{fig:9}
\end{figure}

\begin{figure}[!htbp]
\centering
\includegraphics[width=\columnwidth]{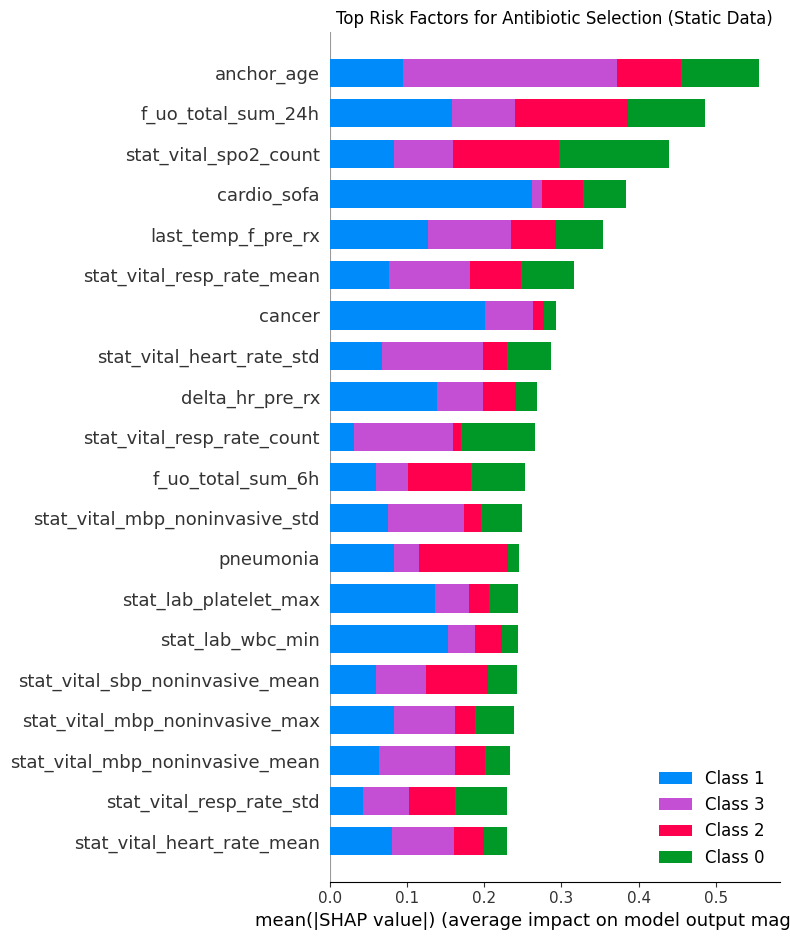}
\caption{Feature Importance for Antibiotic Selection}
\label{fig:10}
\end{figure}

\section{Discussion}
\textbf{The Failure of SepsisFusionFormer:} The underperformance of the Deep Fusion architecture is a significant finding. Theoretically, Gated Attention should capture complex interactions (e.g., "High Temp" attending to "Pneumonia" in text). However, we posit that the "curse of dimensionality" overwhelmed the signal. With only $\sim$2,100 samples in the antibiotic cohort, the network likely memorized noise rather than learning generalized cross-modal attention maps. This suggests that SepsisFusionFormer could be highly effective for more common conditions (e.g., Hypertension) where $N > 100,000$, but for specialized ICU tasks, it is ill-suited.

\textbf{The Power of Orthogonal Experts:} The success of \textbf{SepsisLateFusion} relies on the orthogonality of its experts. The "Monitor" (GRU) observes trends the "Historian" (CatBoost) ignores, while the "Reader" (BioBERT) catches context missed by both. Stacking these diverse probability distributions allows the Meta-Learner to calibrate trust dynamically, resulting in a system that is robust to missing modalities.

\textbf{Vision vs. Text:} The marginal gain from the Vision modality (0.003 AUC) implies high information redundancy. A radiologist's text report ("Reader") likely condenses the salient features of the X-Ray ("Visionary") effectively. This supports a "leaner" deployment strategy where heavy image processing can be omitted in favor of NLP without significant performance loss.

\textbf{Limitations:} First, this study is retrospective and limited to the MIMIC-IV cohort. Second, while we rigorously redacted drug names, the inclusion of pathogen names as a proxy for rapid diagnostics assumes a workflow availability that varies by institution. Finally, the "ground truth" for antibiotic selection represents provider behavior, not necessarily optimal outcome.

\section{Future Work}

\subsection{From Imitation to Optimization: The Causal-RL Frontier}
The most transformative direction for this work lies in the transition from \textbf{Supervised Learning} to \textbf{Optimization}. Currently, the \textbf{SepsisLateFusion} architecture operates as a high-fidelity \textit{behavioral cloning} system, predicting the action a clinician \textit{did} take. The next iteration aims to determine the action a clinician \textit{should} take to maximize patient survival.

To bridge this gap, we propose integrating \textbf{Causal Inference} (e.g., Causal Effect Variational Autoencoders) to estimate Individual Treatment Effects (ITE) for each antibiotic choice. These causal estimates will act as a de-confounded "ground truth" reward signal for an \textbf{Offline Reinforcement Learning (RL)} agent. By training the RL agent on these causal rewards rather than raw outcomes, we can learn a policy that optimizes for survival while penalizing the accumulation of resistance-driving broad-spectrum days, effectively moving from static prediction to evolutionary prescriptive intelligence.

\subsection{Federated Expansion: Scale, Validation, and Fairness}
While the \textbf{SepsisFusionFormer} (Deep Fusion) architecture struggled with the sparsity of the MIMIC-IV antibiotic cohort ($N \approx 2,100$), it remains a theoretically superior architecture for modeling complex, non-linear cross-modal dependencies. To definitively validate this hypothesis, future work will leverage \textbf{Federated Learning} to train across multi-center datasets, such as the eICU Collaborative Research Database and AmsterdamUMCdb.

This expansion serves two critical functions. First, achieving a sample size of $N > 100,000$ will determine if the ``attention starvation'' observed in our deep fusion experiments can be overcome by scale. Second, and more importantly, this allows for rigorous \textbf{Out-of-Distribution (OOD) testing and bias auditing}. By evaluating the ``lean'' \textbf{SepsisLateFusion} ensemble on diverse patient populations with distinct stewardship protocols and demographic profiles, we can quantify and mitigate algorithmic bias, ensuring that the model's high performance is not an artifact of the specific care patterns at Beth Israel Deaconess Medical Center.

\section{Conclusion}
This work presents \textbf{SepsisSuite}, a unified framework that advances the frontier of sepsis AI from passive risk stratification to active therapeutic guidance. Through a rigorous comparative analysis, we challenged the prevailing assumption that end-to-end Deep Learning is inherently superior for electronic health records. Our experiments demonstrated that while the theoretical capacity of the \textbf{SepsisFusionFormer} (Deep Fusion) is vast, it is fundamentally constrained by data sparsity in high-stakes cohorts, leading to "attention starvation" and overfitting.

In contrast, our pivotal transition to a \textbf{Context-Aware Mixture-of-Experts} architecture (SepsisLateFusion) established a new paradigm for data-efficient modeling. By treating modalities as orthogonal experts—the Historian, the Monitor, and the Reader—we achieved State-of-the-Art performance in early detection (\textbf{0.915 AUC at 4 hours pre-onset}) and mortality prediction (\textbf{0.91 AUC}), while setting a novel benchmark for multi-class empiric antibiotic selection (\textbf{0.72 AUC}).

Ultimately, this project validates that clinical utility is best served not by maximizing architectural complexity, but by enforcing architectural interpretability and temporal rigor. The success of the "lean" Trimodal/Quad-modal ensembles provides a reproducible blueprint for the next generation of medical AI: systems that are robust to missing data, transparent in their routing logic, and capable of supporting the complex, granular decisions that define critical care medicine.

\begin{acks}
  We thank the PhysioNet team for the MIMIC-IV dataset access.
\end{acks}

\nocite{*} 

\bibliographystyle{ACM-Reference-Format}
\bibliography{references} 

\end{document}